\documentclass{article}

\usepackage[numbers,square]{natbib}

\usepackage[preprint]{2026}

\usepackage[utf8]{inputenc} 
\usepackage[T1]{fontenc}    
\usepackage{hyperref}       
\usepackage{url}            
\usepackage{booktabs}       
\usepackage{amsfonts}       
\usepackage{nicefrac}       
\usepackage{microtype}      
\usepackage{xcolor}         
\usepackage{soul}
\usepackage{amsmath}
\usepackage{graphicx}
\usepackage{longtable}
\usepackage{booktabs}
\usepackage{caption}
\usepackage{comment}
\usepackage{wrapfig}
\usepackage{graphicx}
\usepackage{subcaption}
\title{Prompting Diffusion Models\\for Zero-Shot Instance Segmentation}

\author{
  İrem Zeynep Alagöz$^{1}$\thanks{Equal contribution.}
  \And
  Nils Morbitzer$^{1,2}$\footnotemark[1]
  \And
  Andrea Ramazzina$^{1,3}$
  \And
  Nassir Navab$^{1,2}$
  \And
  Federico Tombari$^{1,2}$\quad
  Stefano Gasperini$^{1,2,4}$\\\\
  $^1$Technical University of Munich (TUM)\quad
  $^2$Munich Center of Machine Learning (MCML)\\
  $^3$Mercedes-Benz AG\quad
  $^4$Visualais\quad
}

\def\pname{Prompt2Seg}

\begin{document}

\maketitle

\begin{figure}[!h]
  \centering
   \includegraphics[width=\linewidth]{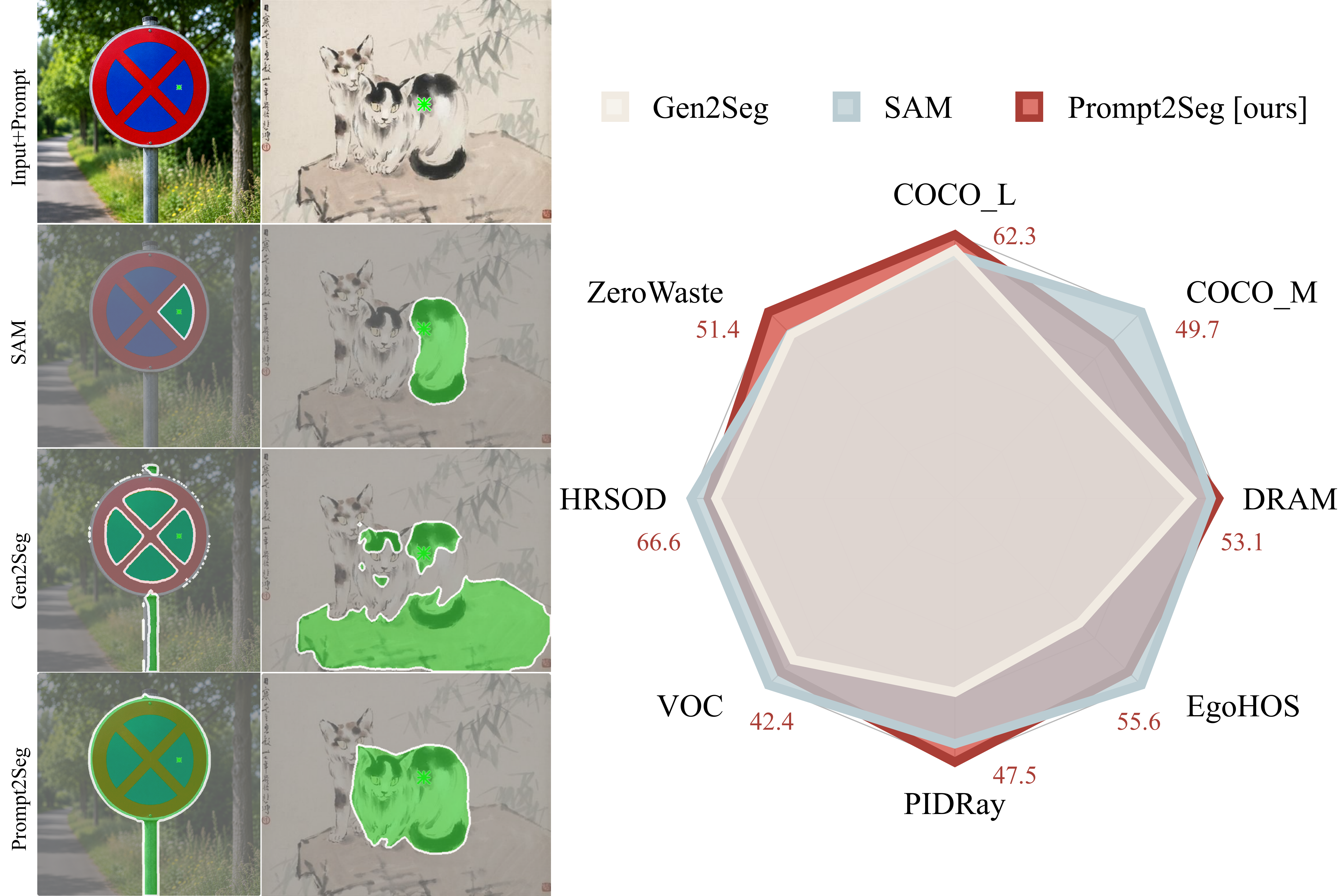}
   \caption{While discriminative-based models~\cite{kirillov2023sam} and prior work leveraging diffusion priors~\cite{khangaonkar2025gen2seg} struggle to segment objects with strong pixel gradients, leading to over-segmentation, our proposed \textbf{Prompt2Seg conditions the diffusion process on the user input} using spatial prompts and delivers semantic-aware, interactive segmentation. The scores in the plot are from our method.}
   \label{fig:teaser}
\end{figure}

\begin{abstract}
Several disruptive research directions have recently emerged in computer vision, including foundation models achieving previously unseen zero-shot performance in scene understanding, even interactively, and generative models that synthesize extremely realistic images. The latter have also been shown to be highly effective in scene understanding tasks thanks to their rich priors. However, for promptable segmentation, foundation models struggle with accurately segmenting an object's region, leading to false positives and over-segmentation. Notably, early attempts that leverage generative priors use prompts only during post-processing, yielding suboptimal segments because the process is agnostic to the user input. In this paper, we target these limitations with \pname{}, a spatial conditioning framework for diffusion-based segmentation. \pname{} augments a frozen diffusion segmentation model with a conditioning branch. Our approach takes spatial prompts, represented as 2D Gaussians or confidence maps, as explicit input signals, training the model to respond directly to user intent. Fine-tuned on a deliberately constrained set of object categories drawn from Hypersim and Virtual KITTI 2, \pname{} generalizes zero-shot to a wide range of unseen object types and visual domains. We evaluate on seven datasets ranging from standard benchmarks to more challenging domains, including paintings, egocentric views, and X-ray data. Furthermore, we demonstrate that \pname{} consistently outperforms the underlying diffusion segmentation backbone across all benchmarks. Our results suggest that the rich priors encoded in generative pretraining, combined with principled spatial conditioning, offer a compelling path toward broadly generalizing interactive segmentation without large-scale mask supervision.
\end{abstract}

\section{Introduction}
In recent years, scene understanding has shifted from models trained and evaluated on fixed, domain-specific datasets toward foundation models, systems capable of generalizing across contexts they have never explicitly seen. One prominent example is SAM~\cite{kirillov2023sam}, which demonstrated pioneering zero-shot performance on instance segmentation and popularized promptable segmentation.

Despite their impressive zero-shot generalization capabilities and high impact in many applications, discriminative models such as SAM are trained purely on label boundaries and therefore lack structural priors about object appearance. As shown in \autoref{fig:teaser}, this becomes particularly apparent for objects with sharp, well-defined internal edges or those without clear boundaries. In these cases, understanding the object structure and semantics, rather than just focusing on the pixel contrast, is critical to produce meaningful segments instead of over-segmentation.

On the other hand, diffusion models, such as Stable Diffusion~\cite{rombach2022stablediffusion}, have been trained to recover images in a self-supervised manner, thereby developing an understanding of object "wholeness" and building up a structural prior. Due to this strong internal representation, diffusion models, originally trained for image synthesis, have been effectively repurposed toward several downstream tasks, such as estimating geometric attributes~\cite{ke2024marigold, fu2024geowizard, ye2024stablenormal}, pose~\cite{gong2023diffpose, wang2023posediffusion}, or motion~\cite{luo2024flowdiffuser}. 

Most recently, diffusion models have also been adapted to segmentation. Here, the few current generative segmenters are either conditioned only on language~\cite{deng2025llamaseg, chen2025gs} or treat user clicks as an afterthought~\cite{khangaonkar2025gen2seg}, which makes them less "controllable" than SAM. More specifically, Gen2Seg~\cite{khangaonkar2025gen2seg} uses its diffusion-based instance features as a static "bank" and utilizes clicks for a post-hoc similarity filter to simulate interactive instance segmentation. However, this pioneering procedure fails to utilize the user's intent to guide the feature extraction process. As shown in \autoref{fig:teaser}, this leads to suboptimal results.

In this paper, we tackle both problems by introducing \pname{}, a finetuning protocol to actively inject spatial prompts to recondition the layers of a frozen diffusion segmentation model. Our model takes these spatial prompts, represented as 2D Gaussians or dense confidence maps, as a direct input signal instead of a post-hoc step. Our method yields the first diffusion model for interactive instance segmentation with strong zero-shot performance.
Additionally, \pname{} reduces the need to collect diverse training data, as we show that it can learn effectively on synthetic images alone and generalize well to real inputs.
Our contributions can be summarized as follows:
\begin{itemize}
    \item For the first time, we show how to adapt diffusion for interactive segmentation using spatial prompts.
    \item We present \pname{}, enabling the conditioning of semantic-aware interactive instance segmentation via diffusion using spatial prompts, represented as 2D Gaussians or confidence maps.
    \item We show that our finetuning strategy yields a strong zero-shot interactive segmentation model, especially on challenging domains, including artistic images (DRAM), X-ray data (PIDRay), and cluttered scenes (ZeroWaste).
\end{itemize}
\section{Related Work}
\subsection{Generative Models for Perceptual Tasks}

Generative modeling has increasingly influenced visual perception, shifting from auxiliary training objectives toward direct inference mechanisms. Early works~\cite{chen2020igpt, he2022mae} show that generative objectives, i.e. autoregressive prediction or masked reconstruction, are effective for learning transferable visual representations from unlabeled data. Subsequent works demonstrated that large-scale text-conditional generation models encode rich, reusable perceptual knowledge as a byproduct of learning to synthesize images from language: features extracted from generatively pretrained models transfer surprisingly well to discriminative tasks, often rivaling representations learned with explicit supervision~\cite{baranchuk2021ddpmsemseg, zhao2023vpd}.

More recent approaches go a step further, repurposing generative models as direct inference engines for dense prediction; rather than extracting features, they generate the output itself through diffusion or autoregressive decoding. This paradigm has proven broadly applicable, spanning geometric estimation~\cite{ke2024marigold, fu2024geowizard, ye2024stablenormal}, pose~\cite{gong2023diffpose, wang2023posediffusion} motion~\cite{luo2024flowdiffuser}, and, most recently, semantic tasks such as segmentation~\cite{khangaonkar2025gen2seg, deng2025llamaseg}. Our work follows this direction, investigating whether the conditioning machinery of diffusion models can be repurposed for promptable instance segmentation, combining the inference-time flexibility of generative decoding with the broad visual understanding acquired during large-scale pretraining.

\subsection{Diffusion Models for Segmentation}
Diffusion models~\cite{ho2020ddpm} have become the de-facto standard for high-quality image synthesis, achieving impressive performance in text-to-image synthesis~\cite{rombach2022stablediffusion} and controlled generation~\cite{zhang2023controlnet}. 
Beyond generation, the internal features of denoising U-Nets have proven surprisingly useful for perception. Baranchuk et al.~\cite{baranchuk2021ddpmsemseg} show that DDPM activations capture semantic information, thereby, enabling few-shot segmentation without any task-specific training. Since then, diffusion models have been explored as visual backbones for a range of perception tasks, with segmentation being among the more recent directions to gain traction \cite{khangaonkar2025gen2seg, deng2025llamaseg, chen2025gs}. Existing approaches can be grouped in four categories: (1) Utilizing diffusion model's strong frozen features for discriminative models \cite{zhao2023vpd, zhu2024vdit}; (2) generating pseudo labels to supervise discriminative models~\cite{nguyen2023datasetdiffusion, ma2023diffusionseg, wu2023diffumask}; (3) inversion-based approaches that repurpose internal attention maps for zero-shot object localization~\cite{yang2024diffpng, liu2024vgdiffzero, ni2023refdiff}, (4) or directly synthesizing segmentation masks in label~\cite{chen2025gs} or RGB space~\cite{khangaonkar2025gen2seg}.

Gen2Seg~\cite{khangaonkar2025gen2seg} shows that diffusion models trained on billions of images serve as strong priors for segmentation, enabling remarkable zero-shot generalization and label-efficient fine-tuning. However, they treat the model as a passive predictor, yet diffusion models are by nature interactive trained to respond to rich multimodal conditioning. We show that their conditioning mechanism can be repurposed for user-guided instance segmentation. This yields a promptable segmentation framework that inherits the strong generalization of generative pretraining.

\begin{figure}
  \centering
   \includegraphics[width=\linewidth]{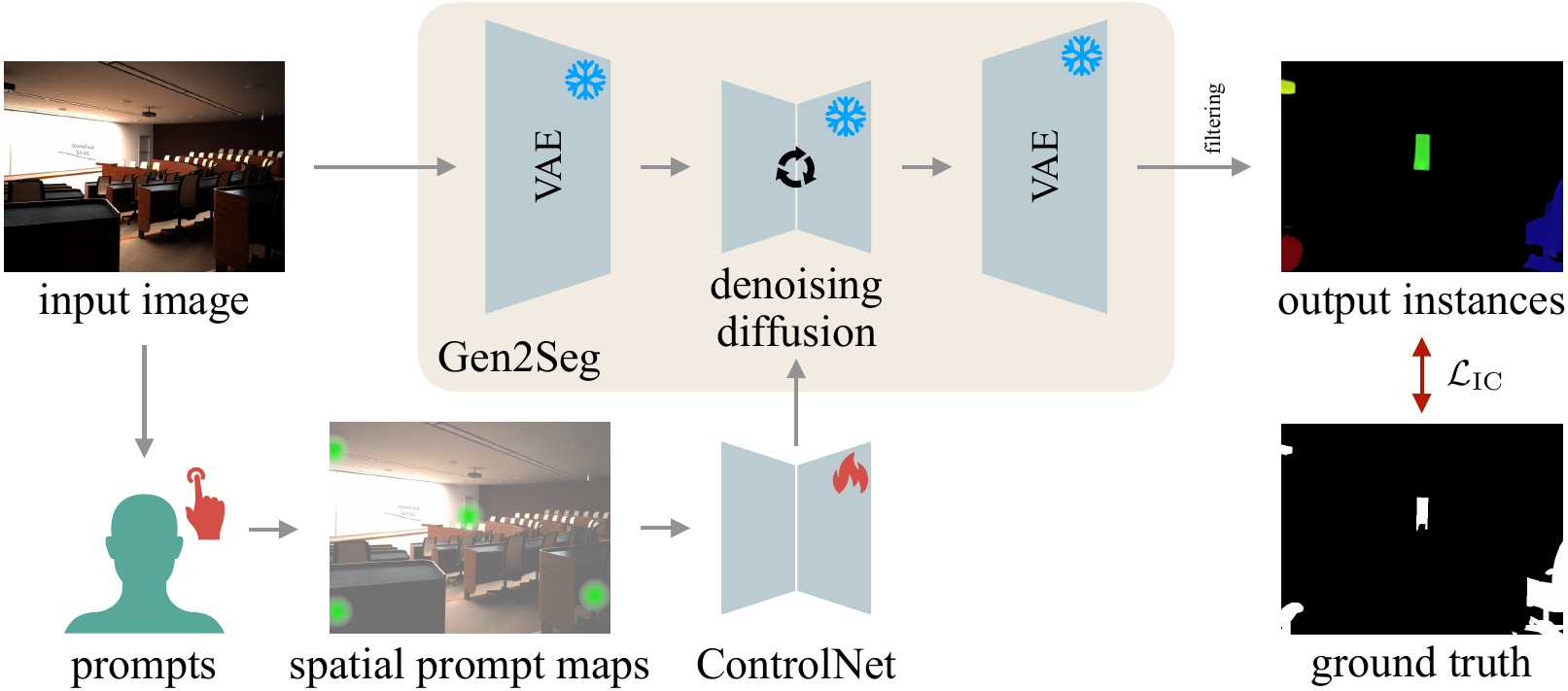}
   \caption{\textbf{Our Prompt2Seg framework}. Prompt2Seg segments any instance by guiding the diffusion process with a ControlNet conditioned on the user prompts as Gaussian maps.}
   \label{fig:framework}
\end{figure}

\subsection{Promptable Instance Segmentation}

Interactive segmentation has been extensively studied through discriminative approaches, where models directly predict segmentation masks given user-provided prompts. Click-based methods such as SimpleClick~\cite{liu2023simpleclick} leverage strong vision encoders to produce accurate masks from sparse point inputs, while the Segment Anything Model (SAM)~\cite{kirillov2023sam} and its successor SAM 2~\cite{ravi2025sam2} scale this paradigm to large-scale data, achieving class-agnostic segmentation from clicks, boxes, or masks. To incorporate semantic understanding, Grounded SAM~\cite{ren2024groundedsam} and SAM 3~\cite{carion2026sam3} couple open-vocabulary detectors with SAM, enabling language-guided segmentation. Despite their strong performance, discriminative models require task-specific architectures and large amounts of annotated data to learn each new prompt modality. Generative models, by contrast, intrinsically develop rich visual representations during pretraining that can be unlocked with only lightweight adaptation~\cite{gabeur2026visionbanana}, suggesting that a generative backbone is inherently better suited for few-shot generalization to new prompting conditions. Recent work has begun to exploit this property for segmentation: GS~\cite{chen2025gs} reformulates language-driven segmentation as a label diffusion process, and LlamaSeg~\cite{deng2025llamaseg} treats it as autoregressive mask token generation conditioned on language instructions. Most closely related to our work, Gen2Seg~\cite{khangaonkar2025gen2seg} finetunes Stable Diffusion~\cite{rombach2022stablediffusion} for category-agnostic instance segmentation by casting it as an image-to-image translation problem, where instances are distinguished by color. However, its prompting mechanism reduces to post-hoc color-based instance retrieval; that is the model is never trained to incorporate spatial user prompts, which we show leads to consistently inferior performance across all zero-shot benchmarks (see Table~\ref{tab:quantitative_zero_shot_1}). More broadly, existing generative approaches either rely exclusively on language conditioning or, like Gen2Seg, treat spatial interaction as an auxiliary mechanism. In contrast, our approach explicitly trains the model to condition on spatial prompts; thereby, integrating interactivity as a core component of the generative process and leveraging generative pretraining's few-shot adaptability for interactive instance segmentation.


\section{Method}
\label{sec:method}

We introduce \pname{}, a spatially prompted diffusion model for category-agnostic interactive instance segmentation. Given an image and a sparse spatial prompt, \pname{} predicts the mask of the prompted object. The central idea is to preserve the visual priors of a pretrained diffusion model while adding an explicit spatial control interface to guide the segmentation. To this end, we train only a lightweight conditioning branch that injects prompt-dependent residuals into the frozen backbone.

Let $x \in \mathbb{R}^{3\times H\times W}$ be an input image and let $c \in \mathbb{R}^{C\times H\times W}$ be a spatial condition derived from the sparse user prompts. \pname{} predicts an RGB instance-coloring map
\begin{equation}
    y = f_{\theta,\phi}(x,c),
    \qquad
    y \in \mathbb{R}^{3\times H\times W},
\end{equation}
where $\theta$ denotes the frozen diffusion segmentation backbone and $\phi$ denotes the trainable prompt-conditioning branch. 

\subsection{Pretrained Diffusion Segmentation Backbone}
\label{sec:diffusion_backbone}

We build on Gen2Seg~\cite{khangaonkar2025gen2seg}, which repurposes a Stable-Diffusion-initialized image-to-image model for instance segmentation. Gen2Seg predicts an RGB instance-coloring map for the entire image: pixels belonging to the same region are encouraged to share a color, different regions are encouraged to be separated in color space, and the background is mapped to black. A target instance is then recovered only after prediction by post-processing this full-image representation, e.g., by thresholding distances in the predicted color space. This can introduce ambiguities when nearby instances receive similar colors. In contrast, \pname{} keeps the same RGB output representation but changes its semantics: the prompt selects the target instance before decoding, so the model directly predicts a single prompt-conditioned foreground region against a black background. Thus, \pname{} avoids full-image instance selection at post-processing time while leaving the pretrained diffusion segmentation backbone and output loss unchanged.

\subsection{Spatial Prompt Representation}
\label{sec:prompt_encoding}

A point prompt is a pixel coordinate within the pixels of the target object. We transform this point into an image-like continuous spatial map that can be consumed by a convolutional conditioning branch. Let $\mathcal{P}^{+}=\{r_i\}_{i=1}^{P}$ be the set of positive prompts, where $r_i=(u_i,v_i)$ is an image coordinate inside the target object. For a pixel coordinate $q$, we define a Gaussian prompt map
\begin{equation}
    G_{\sigma}(q;r_i)
    =
    \exp\!\left(
        -\frac{\|q-r_i\|_2^2}{2\sigma^2}
    \right),
\end{equation}
where $\sigma$ controls the spatial extent of the prompt. Multiple positive prompts are aggregated by a pixel-wise maximum,
\begin{equation}
    h_{\sigma}^{+}(q)
    =
    \max_{r_i\in\mathcal{P}^{+}} G_{\sigma}(q;r_i).
\end{equation}
This aggregation keeps the condition's dimensionality fixed while allowing multiple clicks.

Our main model uses a two-scale Gaussian prompt encoding. We construct
\begin{equation}
    h_s^{+}(q)=h_{\sigma_s}^{+}(q),
    \qquad
    h_l^{+}(q)=h_{\sigma_l}^{+}(q),
    \qquad
    \sigma_s < \sigma_l.
\end{equation}

The smaller bandwidth $\sigma_s$ preserves precise localization at the click, whereas the larger bandwidth $\sigma_l$ increases the spatial support of the prompt signal. This two-scale encoding, therefore, provides both point-level and neighborhood-level cues to the prompt branch. The final spatial condition is a three-channel tensor
\begin{equation}
    c = \Psi(h_s^{+},h_s^{+},h_l^{+}) \in \mathbb{R}^{3\times H\times W},
\end{equation}
where $\Psi$ maps the two-scale prompt representation to the channel layout expected by the conditioning branch. The prompt is provided as an explicit spatial signal before prediction, rather than being used as a post-processing query on an already generated instance map, as in Gen2Seg. By doing so, our model learns to directly focus on generating of the prompted object segmentation.

While the description above focuses on a 2D Gaussian representation, our prompt can differ from the sparse clicks paradigm. This is useful when the region of interest is produced by another model (e.g., a dense confidence or uncertainty map) rather than by a user.

\subsection{Prompt Injection with a ControlNet Branch}
\label{sec:controlnet_prompt_injection}

We inject spatial prompts through a ControlNet-style branch~\cite{zhang2023controlnet}. 
Since Gen2Seg uses a single-step formulation, all predictions are evaluated at a fixed timestep $t^\star$. 
Let $z_x$ denote the VAE latent of the input image, $h$ the encoder hidden states, and $c$ the spatial conditioning input. 
The trainable ControlNet branch predicts down-block and mid-block residuals,
\begin{equation}
    \left(
    \Delta^{(1)},\ldots,\Delta^{(L)},\Delta^{\mathrm{mid}}
    \right)
    =
    g_{\phi}(z_x,t^\star,h,c).
\end{equation}
These residuals are injected into the frozen Gen2Seg U-Net via its additional residual inputs:
\begin{equation}
    \hat{u}
    =
    f_{\theta}
    \left(
        z_x,t^\star,h;
        \Delta^{(1)},\ldots,\Delta^{(L)},\Delta^{\mathrm{mid}}
    \right),
\end{equation}
where $\hat{u}$ is the diffusion-model prediction. 
The Gen2Seg scheduler maps this prediction to a segmentation latent,
\begin{equation}
    \hat{z}_y = \mathcal{S}_{t^\star}(z_x,\hat{u}),
    \qquad
    \hat{y} = D_{\mathrm{VAE}}(\hat{z}_y),
\end{equation}
where $\hat{y}$ is the decoded RGB instance-coloring map.

During training, the VAE, text encoder, and denoising U-Net remain frozen, while the ControlNet parameters $\phi$ are optimized.

\subsection{Prompt-Conditioned Supervision}
\label{sec:training}

We inherit the instance-coloring supervision of Gen2Seg. For a set of supervised regions $\mathcal{S}$, the Gen2Seg objective encourages pixels within each region to share a color, encourages different regions to be separated in color space, and fixes the background mean to black. We use the original objective
\begin{equation}
    \mathcal{L}_{\mathrm{IC}}
    =
    \mathcal{L}_{\mathrm{var}}
    +
    \lambda_{\mathrm{sep}}\mathcal{L}_{\mathrm{sep}}
    +
    \lambda_{\mathrm{mean}}\mathcal{L}_{\mathrm{mean}},
\end{equation}
with the same definitions and hyperparameters as Gen2Seg. We include the full loss definition in the appendix for completeness.

Instead of coloring all instances in an image~\cite{khangaonkar2025gen2seg}, \pname{} is trained to predict only the instance specified by the spatial prompt. For each training image, we sample one or more target instances, construct the spatial condition from prompts inside those instances, and keep only the selected instances as foreground in the target instance-coloring map. All non-target pixels are assigned to the black background. Thus, the supervision matches the interactive task: the correct output is not an all-instance segmentation, but the mask of the prompted target.

This prompt-conditioned target construction is category-agnostic. The model receives no semantic class labels, and the prompt tensor contains only spatial information. Therefore, the trainable branch must learn to associate the prompt location with image evidence and object boundaries rather than memorizing a fixed label space. This design allows us to train on a deliberately constrained set of synthetic categories and evaluate whether the frozen diffusion prior generalizes to unseen categories and domains.

\subsection{Inference}
\label{sec:inference}

At inference time, the model receives an image and a spatial condition, constructed from the user point prompts or from the confidence or uncertainty map.

The final binary mask is obtained by thresholding the distance from black (i.e., background) in RGB embedding space:
\begin{equation}
    \hat{m}(q)
    =
    \mathbf{1}
    \left[
    \|y_q-\mathbf{0}\|_2 > \tau
    \right],
\end{equation}
where $\tau$ is selected on a validation set and kept fixed during evaluation. In the interactive setting, $\hat{m}$ is the mask of the prompted target object.

\section{Experiments and Results}
\label{sec:experimental_results}

\paragraph{Datasets \& Metrics}
\textbf{Training:} Following Gen2Seg~\cite{khangaonkar2025gen2seg}, we used only two synthetic datasets for fine-tuning: Hypersim~\cite{roberts2021hypersim} and Virtual KITTI 2~\cite{cabon2020virtualkitti}. Hypersim covers a wide range of indoor environments, bathrooms, bedrooms, libraries, kitchens, and more, while Virtual Kitti 2 provides photorealistic outdoor driving scenes. We also followed their filtering procedure.
Despite their photorealistic quality and scene variety, both datasets' combined category coverage is inherently limited by their domain focus, lacking annotations for many common everyday object categories (e.g., animals, people). These characteristics make Gen2Seg's training set a deliberately constrained testbed to cleanly isolate and evaluate our approach's zero-shot generalization capacity to object categories and visual domains entirely absent from training.
\textbf{Evaluation:} Once fine-tuned on the limited category set described above, we evaluate zero-shot generalization for interactive instance segmentation using mIoU on seven benchmarks spanning diverse domains and object types. Following Gen2Seg~\cite{khangaonkar2025gen2seg} and SAM~\cite{kirillov2023sam}, we adopt four benchmarks, as a direct basis for comparison: COCO~\cite{lin2014coco} (validation set, but categories seen during fine-tuning are excluded), DRAM~\cite{cohen2022dram} (artistic imagery), EgoHOS~\cite{zhang2022egohos} (egocentric scenes), and PIDRay~\cite{zhang2023pidray} (luggage X-rays). We further include Pascal VOC~\cite{everingham2010voc} and HRSOD~\cite{zeng2019hrsod} to broaden evaluation coverage to natural images with diverse object categories and high-resolution salient objects, respectively. Together, these benchmarks span a wide range of visual styles and object categories absent from our training data, providing a rigorous test of generalization. We follow the evaluation splits used by Gen2Seg for DRAM and COCO. For EgoHOS, we report results on both the in-domain and out-of-domain test splits. For the remaining datasets, we use the standard evaluation splits: VOC2012 val, HRSOD val, ZeroWaste test, and the PIDRay easy test.

\paragraph{Baselines}
We compare against three main baselines. Gen2Seg~\cite{khangaonkar2025gen2seg}, most closely related to our work, is a generative approach that fine-tunes Stable Diffusion for category-agnostic instance segmentation by casting it as an image-to-image translation problem, where instances are distinguished by color. Our method directly extends this framework by introducing principled spatial prompting, allowing us to isolate the contribution of explicit prompt conditioning over Gen2Seg's post-hoc color-based instance retrieval. Beyond Gen2Seg's primary Stable Diffusion-based model, we also report results for its MAE and DINO-based variants.
SimpleClick~\cite{liu2023simpleclick} is a state-of-the-art promptable segmenter built on a MAE-pretrained ViTDet backbone. To ensure a fair comparison, we fine-tune it on the same synthetic dataset as ours using its official training code, following the protocol of Gen2Seg. This makes SimpleClick a natural discriminative baseline for assessing whether standard architectures can generalize beyond their supervised categories. SAM~\cite{kirillov2023sam} is used off-the-shelf with its ViT-H backbone as an upper-bound reference, having been supervised on SA-1B~\cite{kirillov2023sam} with over one billion annotated masks.

\paragraph{Implementation Details}
The implementation details are provided in the supplementary material.

\begin{table*}[!t]
\caption{\textbf{Zero-shot quantitative evaluation} on 7 different datasets. Diffusion-based models are reported in the lower block. $\dagger$: ViT-H model trained on SA-1B~\cite{kirillov2023sam}, unlike the others trained only on synthetic data from \cite{roberts2021hypersim, cabon2020virtualkitti}. P2Seg is short for \pname{}.}
\setlength{\tabcolsep}{3.8pt}
\centering
\begin{tabular}{lcccccccccccc}
\toprule
&&COCO&&&&&&& \\
\cmidrule(lr){2-4}
Method & $S$ & $M$ & $L$ & DRAM & EgoHOS & PIDRay & VOC & HRSOD & ZeroWaste\\
\midrule
SAM \textbf{$\dagger$} & 57.0 & 59.7 & 57.6 & 51.6 & 59.8 & 44.2 & 45.5 & 71.2 & 45.4 \\
SimpleClick  & 34.6 & 45.2 & 56.5 & 48.7 & 51.8 & 14.1 & 42.3   & 74.2 & 45.3\\
\midrule
MAE-H & 4.3 & 27.2 & 59.9 & 48.7 & 33.3 & 25.4 & 37.0 & 66.4 & 36.9 \\
Gen2Seg & 8.3 & 38.4 & 58.7 & 47.5 & 40.0 & 34.9 & 39.4 & 64.5 & 45.2\\


\textbf{P2Seg}-Concat
& 4.3 & 26.1 & 53.7 
& 47.0 & 37.3 & 24.0 & 32.1 & 53.5 & 39.4\\

\textbf{P2Seg}-Adapter
& 6.9 & 37.2 & 60.4 
& 48.3 & 46.2 & 41.7 & 39.0 & 64.4 & 45.3 \\

\textbf{P2Seg} & \textbf{12.1} & \textbf{49.7} & \textbf{62.3} & \textbf{53.1} & \textbf{55.6} & \textbf{47.5} & \textbf{42.4} & \textbf{66.6} & \textbf{51.4}\\

\bottomrule
\end{tabular}
\label{tab:quantitative_zero_shot_1}
\end{table*}

\subsection{Results}

\subsubsection{Zero-Shot Evaluation}
Table~\ref{tab:quantitative_zero_shot_1} reports zero-shot performance on seven benchmarks. Following Gen2Seg~\cite{khangaonkar2025gen2seg}, we additionally split COCO~\cite{lin2014coco} by instance size to examine how performance varies with object scale.

Prompt2Seg consistently outperforms SimpleClick~\cite{liu2023simpleclick} and all Gen2Seg variants across all datasets, confirming that neither discriminative pretraining nor post-hoc color-based prompting generalizes as effectively as our approach. Compared to SAM~\cite{kirillov2023sam}, trained on the large-scale, human-annotated SA-1B dataset, the picture is more nuanced. SAM performs better on benchmarks that closely resemble its training distribution, namely COCO~\cite{lin2014coco}, EgoHOS~\cite{zhang2022egohos}, Pascal VOC~\cite{everingham2010voc}, and HRSOD~\cite{zeng2019hrsod}. In contrast, Prompt2Seg surpasses SAM on DRAM~\cite{cohen2022dram}, PIDRay~\cite{zhang2023pidray}, and ZeroWaste~\cite{bashkirova2022zerowaste}, datasets that are more long-tail in nature, covering artistic imagery, X-ray scans, and cluttered waste scenes with ambiguous object boundaries.
This pattern is consistent with generative priors being more robust to domain shift, having been exposed to a substantially broader visual distribution during pretraining than discriminative models trained on curated segmentation data.
Interestingly, for COCO, Prompt2Seg outperforms SAM on large objects, while SAM gains the upper hand on medium-sized instances. Furthermore, our method exposes substantial limitations on small objects, a weakness shared by Gen2Seg, suggesting it may be a fundamental challenge for diffusion-based approaches rather than specific to our prompting mechanism.

Crucially, despite being fine-tuned solely on synthetic data, Prompt2Seg closes the gap to SAM on EgoHOS, VOC, and HRSOD, where SAM leads by only 3-5 IoU points, while the gap widens substantially on 
$\text{COCO}_S$ and $\text{COCO}_M$. This demonstrates that principled spatial prompting, combined with the rich priors of generative pretraining, offers a compelling and low-effort path to generalizable interactive instance segmentation.

\begin{figure}[t]
  \centering
   \includegraphics[width=\linewidth]{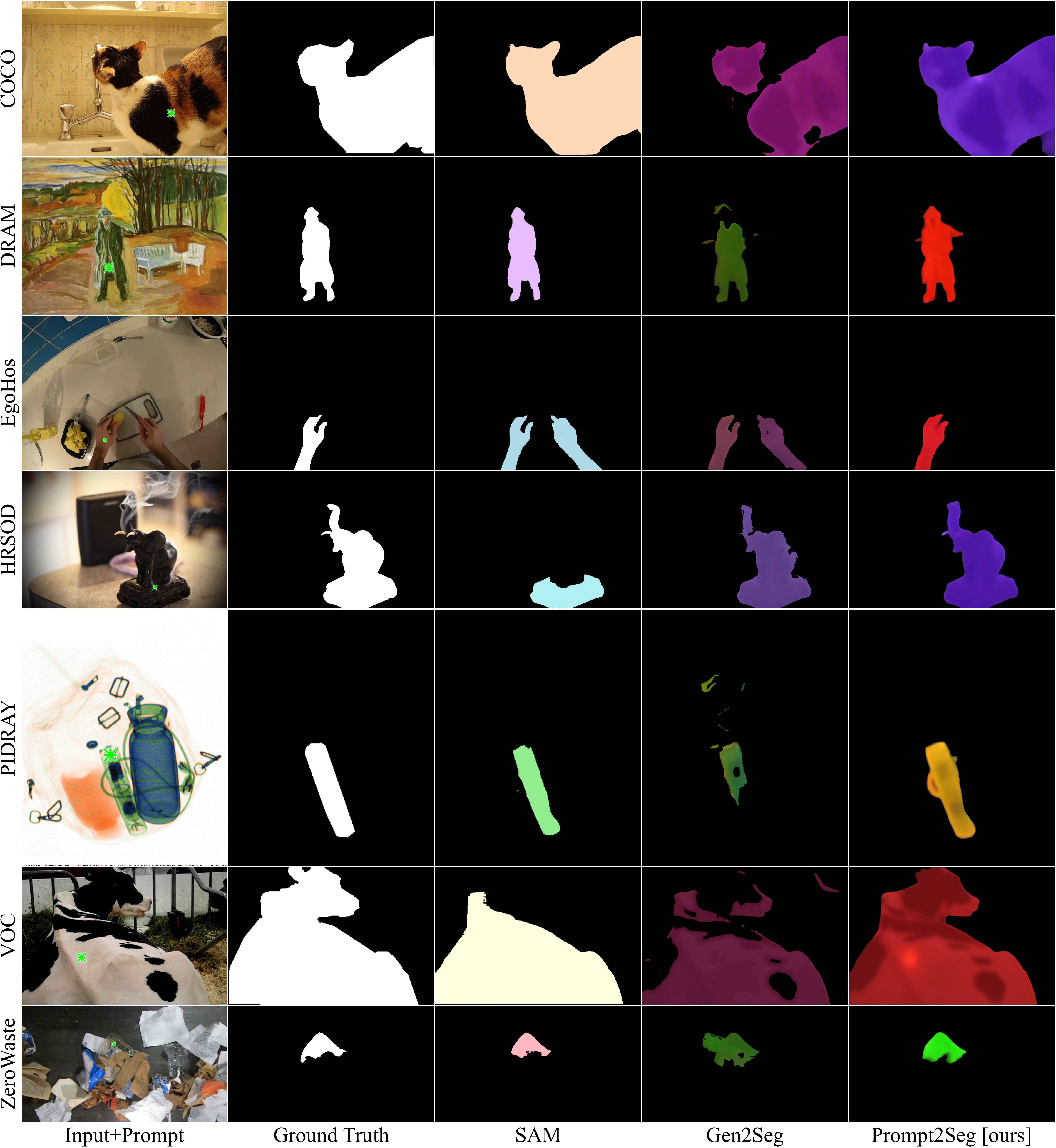}
   \caption{\textbf{Qualitative Results} from various datasets (left) comparing Prompt2Seg with SAM and Gen2Seg. SAM masks are binary, unlike the others, which output features, shown here masked.}
   \label{fig:qualitatives}
\end{figure}

\subsection{Qualitative Results}
\autoref{fig:qualitatives} qualitatively compare Gen2Seg~\cite{khangaonkar2025gen2seg} and SAM~\cite{kirillov2023sam} with our \pname{} across seven datasets. The qualitative results confirm the quantitative findings, suggesting the effectiveness of our introduced prompting mechanism over Gen2Seg's post-selection process.

While Gen2Seg's extracted instance masks look competitive to ours for HRSOD~\cite{zeng2019hrsod}, it tends to underperform on instances with complex or irregular boundaries, such as PIDRay~\cite{zhang2023pidray} or ZeroWaste~\cite{bashkirova2022zerowaste}. In addition, Gen2Seg tends to struggle with objects having contrastive colors. For example, it assigns the cat's (COCO~\cite{lin2014coco}) and cow's (VOC~\cite{everingham2010voc}) black patterns to the background. Interestingly, for EgoHos~\cite{zhang2022egohos}, Gen2Seg and SAM cluster both visible hands together. This could stem from semantic similarity between arms and the ambiguity arising from both arms belonging to the same person. Nevertheless, precise separation of individual arms is often critical in egocentric settings, where our spatial prompting mechanism seems to effectively disambiguate both arms.

\subsection{Beyond Sparse Spatial Prompts}
\label{sec:beyond_sparse_prompts}

\begin{wraptable}{r}{0.4\columnwidth}
\vspace{-1.3cm}
\centering
\caption{\textbf{RoadAnomaly.} P2Seg is applied on RbA or Mask2Former outputs.}
\begin{tabular}{lcc}
\toprule
Method & AP $\uparrow$ & FPR $\downarrow$ \\
\midrule
RbA~\cite{nayal2023rba} 
& 85.42 & 6.92 \\
RbA + \textbf{P2Seg}
& \textbf{85.83} & \textbf{2.25} \\
\midrule
Mask2Former~\cite{cheng2022m2f} 
& 78.45 & 11.83 \\
M2F + \textbf{P2Seg}
& \textbf{88.41} & \textbf{4.95} \\
\bottomrule
\end{tabular}
\label{tab:roadanomaly_conditioning}
\vspace{-0.5cm}
\end{wraptable}

\pname{} is not limited to sparse prompts such as points (e.g., user clicks): it can also use dense maps as spatial conditioning signals, such as the outputs of another model. We evaluate this setting on RoadAnomaly~\cite{lis2019roadanomaly} using RbA~\cite{nayal2023rba}, an anomaly segmentation model, and Mask2Former~\cite{cheng2022m2f}, a closed-set semantic segmentation model. \pname{} is conditioned on RbA anomaly-score maps or on uncertainty maps derived from Mask2Former predictions using the RbA scoring formulation.

Table~\ref{tab:roadanomaly_conditioning} shows that \pname{} improves both source signals, indicating that it does not simply reproduce the conditioning map but refines noisy dense cues into more reliable anomaly masks.

\begin{figure}[t]
  \centering
    \noindent

    \includegraphics[width=\linewidth]{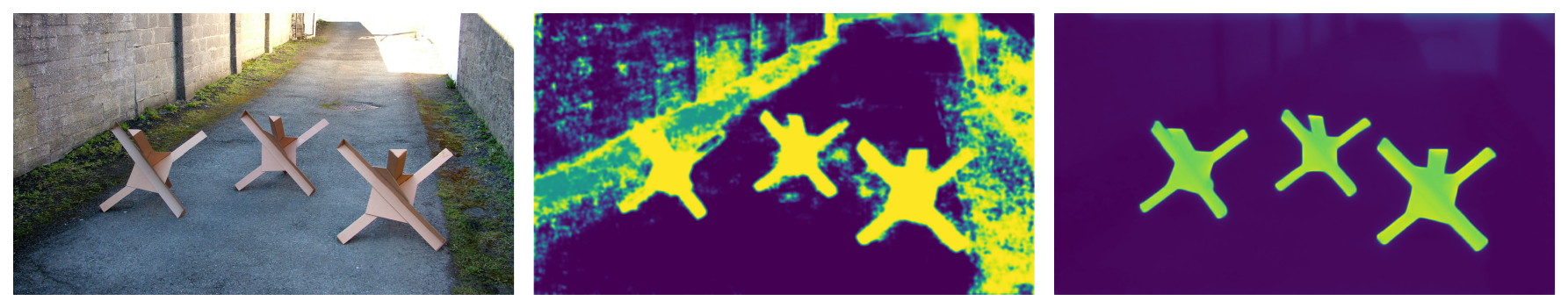}

    \vspace{-4.5pt}
    
  \begin{minipage}[t]{0.32\linewidth}
    \centering\small Input image
  \end{minipage}\hfill
  \begin{minipage}[t]{0.32\linewidth}
    \centering\small Source map
  \end{minipage}\hfill
  \begin{minipage}[t]{0.32\linewidth}
    \centering\small P2Seg output
  \end{minipage}

  \caption{\textbf{Qualitative results on RoadAnomaly~\cite{lis2019roadanomaly}.}
  Conditioned on the source uncertainty map, P2Seg suppresses noisy background responses while preserving the true anomalous objects.}
  \label{fig:qualitatives_ra}
\end{figure}

Figure~\ref{fig:qualitatives_ra} illustrates this behavior qualitatively: the source map captures the anomalous objects but contains substantial background noise, while \pname{} suppresses spurious activations and produces a cleaner, more localized prediction. These results show that \pname{} can serve as a dense-prompt refinement module beyond sparse spatial prompts, opening new opportunities to refine the otherwise jittery and imprecise uncertainty maps.

\begin{wraptable}{r}{0.6\columnwidth}
\setlength{\tabcolsep}{3.8pt}
\vspace{-0.5cm}
\centering
\caption{\textbf{Masks boundary evaluation.} Comparison with Gen2Seg on COCO$_L$.}
\begin{tabular}{lcccc}

\toprule

Method & Boundary F & Precision & Recall & Boundary IoU \\

\midrule

Gen2Seg & 47.09 & 42.62 & \textbf{52.61} & 19.16 \\
\textbf{P2Seg}    & \textbf{52.63} & \textbf{56.43} & 49.31 & \textbf{21.82} \\

\bottomrule

\end{tabular}
\label{tab:boundary}
\end{wraptable}
\subsection{Edge Detection Accuracy}

\autoref{tab:boundary} reports an evaluation of Gen2Seg and ours in terms of the quality of their segmentation boundaries. Despite using all the Gen2Seg model parameters, ours delivers superior mask boundaries by directly conditioning the diffusion process via the spatial prompt.

\section{Limitations and Conclusion}
\label{sec:conclusion}

\paragraph{Limitations and Future Work:} Despite its overall effectiveness, as shown in \autoref{tab:quantitative_zero_shot_1} on $\text{COCO}_S$, our model struggles with small objects. We inherit this issue from Gen2Seg~\cite{khangaonkar2025gen2seg}, which we incorporated frozen and still outperformed on small instances, too. This limitation could be due to the inherent resolution bottleneck of latent diffusion models, where aggressive spatial compression limits the model's ability to represent fine-grained details of small instances~\cite{rombach2022stablediffusion}.
Then, even though \pname{} narrows the gap to discriminative models on natural images and even overtakes SAM in multiple settings, our model still lags behind SAM in some cases due to SAM's large amount of task-specific, annotated training data. Future work could explore how to further leverage the diffusion priors. Moreover, we show failure cases in \autoref{fig:supp_failure_cases}. With a single spatial prompt, the model cannot always understand the desired granularity. Training with multiple positive prompts on the same instance or negative prompts on other objects could help here. Additionally, hierarchical solutions could be explored to output masks at different granularities.

\paragraph{Conclusion:}
We presented the simple and effective \pname{}, a generalizable spatial prompting framework for diffusion-based interactive instance segmentation. \pname{} enables interactive click-based segmentation powered by a diffusion model. With our experiments across seven diverse datasets, we demonstrated that conditioning based on spatial prompting significantly improves over post-hoc selection strategies for diffusion-based models, narrowing the gap and even outperforming strong supervised discriminative models such as SAM~\cite{kirillov2023sam} in multiple settings while relying solely on synthetic training data. Furthermore, we showed how our method can effectively adapt to take as conditioning input an uncertainty map instead of the user's clicks, leading to refined uncertainty masks over OOD objects.

\bibliography{2026}
\bibliographystyle{ieeetr}


\appendix

\clearpage
\section{Implementation Details}
\label{sec:implementation_details}

\paragraph{Backbone and conditioning branch.}
We initialize the segmentation backbone from the official Gen2Seg Stable-Diffusion checkpoint~\cite{khangaonkar2025gen2seg}. The VAE, CLIP text encoder, and denoising U-Net are kept frozen throughout training. We initialize the ControlNet branch from the frozen U-Net weights and train only the ControlNet parameters. All experiments use an empty text prompt, so the model receives task-specific information only through the spatial condition.

\paragraph{Prompt construction.}
For each selected target instance, we sample a positive point from the interior of the ground-truth mask using the maximum of the Euclidean distance transform. This avoids ambiguous boundary clicks during training and provides a stable one-click supervision signal. The prompt is converted into Gaussian heatmaps at two spatial scales. Unless otherwise stated, we use $\sigma_s=8$ and $\sigma_l=16$ pixels, which performed best in our prompt-encoding ablation. Multiple prompts, when present, are combined using a pixelwise maximum. 

\paragraph{Prompt-conditioned targets.}
Training targets are constructed from instance-colored ground-truth masks. For each image, we randomly select valid target instances above a minimum area threshold and set all non-target pixels to black. Thus, the target instance-coloring map contains only the prompted object or objects as foreground, while other annotated objects are treated as background. This converts the original all-instance supervision into a prompt-conditioned target-only prediction task.

\paragraph{Mask extraction.}
The decoded output is an RGB instance-coloring map in the range $[0,255]$. We convert it to a binary mask by thresholding the Euclidean distance of each pixel from black. The threshold is selected with the ablations and then kept fixed for all test datasets. For the reported interactive segmentation experiments, we use the same one-click protocol across all datasets.

\paragraph{Dense-prompt extension.}
For road anomaly experiments, the spatial condition is derived from a frozen uncertainty critic instead of user clicks. We compute the anomaly response from the critic logits, robustly normalize it using image-wise percentiles, smooth the response with local averaging, retain high-response regions, and map the resulting dense condition to $[-1,1]$. The dense map is then provided to the same ControlNet conditioning interface. No change is made to the frozen diffusion backbone or to the instance-coloring objective.

\pagebreak

\section{Ablation Studies}

\subsection{Instance Count}
\begin{table}[h]
\centering
\caption{Ablation on the number of target instances used during training}
\small
\begin{tabular}{lccc}
\toprule
\textbf{\# Instances} & $\text{COCO}_S$ & $\text{COCO}_M$ & $\text{COCO}_L$ \\
\midrule
1  & 9.26	& 43.98	& 64.37 \\
4  & \textbf{9.95}	& \textbf{47.99}	&  64.09 \\
8  & 7.86	&  43.67	&  \textbf{65.04} \\
\bottomrule
\end{tabular}

\label{tab:num_instance_ablation}
\end{table}
Table~\ref{tab:num_instance_ablation} ablates the number of target instances used during training. Using four instances gives the best performance on COCO$_S$ and COCO$_M$, while eight instances slightly improves COCO$_L$ but degrades smaller scales. We therefore use four instances as the default, as it provides the best overall trade-off.

\subsection{Gaussian Size}
\begin{table}[h]
\centering
\caption{Ablation on the standard deviation of the Gaussian point-prompt encoding.}
\small
\begin{tabular}{lccc}
\toprule
\textbf{Gaussian $\sigma_s$ (px)} & $\text{COCO}_S$ & $\text{COCO}_M$ & $\text{COCO}_L$ \\
\midrule
1  & 0.7  & 5.2  & 27.6 \\
8  & \textbf{10.6} & \textbf{43.3} & \textbf{60.7} \\
16 & 9.3  & 41.0 & 60.4 \\
\bottomrule
\end{tabular}

\label{tab:gaussian_sigma_ablation}
\end{table}
\begin{table}[h]
\centering
\caption{Ablation of single- versus dual-scale Gaussian point-prompt encodings. The object-area threshold for training filtering is fixed to 300 pixels for all runs.}
\small
\begin{tabular}{lccc}
\toprule
\textbf{Encoding} & $\text{COCO}_S$ & $\text{COCO}_M$ & $\text{COCO}_L$ \\
\midrule
$\sigma=8$ px & 10.0 & 48.0 & 64.1 \\
$\sigma\in\{8,16\}$ px & 12.1 & 49.7 & 62.3 \\
\bottomrule
\end{tabular}

\label{tab:channel_ablation}
\end{table}

Table~\ref{tab:gaussian_sigma_ablation} ablates the standard deviation of the Gaussian point-prompt encoding. Very narrow prompts provide insufficient spatial context, while overly broad prompts reduce localization precision. A moderate value of $\sigma_s=8$ px performs best across object scales.

Table~\ref{tab:channel_ablation} compares single- and dual-scale Gaussian encodings. Using two scales, $\sigma \in \{8,16\}$ px, improves COCO$_S$ and COCO$_M$, indicating that multi-scale prompt encoding provides a more robust conditioning signal. We use the dual-scale encoding as the default.

\subsection{Area Filtering}
\begin{table}[!h]
\centering
\caption{Ablation on the minimum object-area threshold used for filtering training instances. A threshold of 0 indicates no area filtering.}
\small
\begin{tabular}{lccc}
\toprule
\textbf{px$^2$} & $\text{COCO}_S$ & $\text{COCO}_M$ & $\text{COCO}_L$ \\
\midrule
0   & \textbf{10.6} & 43.3 & 60.7 \\
100 & 10.4 & 44.4 & 59.4 \\
300 & 10.0  & \textbf{48.0} & \textbf{64.1} \\
400 & 10.3 & 42.2 & 58.8 \\
500 & 9.4  & 44.2 & 58.8 \\
\bottomrule
\end{tabular}

\label{tab:area_filtering_ablation}
\end{table}
Table~\ref{tab:area_filtering_ablation} evaluates the minimum object-area threshold used to filter training instances. Since only a few target instances are sampled per image, very small objects often provide weak conditioning signals and are difficult to segment reliably. Filtering instances below 300 px$^2$ improves COCO$_M$ and COCO$_L$ while maintaining competitive COCO$_S$ performance; therefore, we use 300 px$^2$ as the default.

\subsection{Magnitude Thresholding}
\begin{table}[h]
\centering
\caption{Ablation on the magnitude thresholding $\tau$.}
\small
\begin{tabular}{lccc}
\toprule
\textbf{$\tau$} & $\text{COCO}_S$ & $\text{COCO}_M$ & $\text{COCO}_L$ \\
\midrule
50  & 7.7  &	38.8  & 61.3 \\
75  & 9.0  &	43.1 & 63.6 \\
100  & 10.1	& 45.5 &	\textbf{63.3} \\
150  & \textbf{12.1} & \textbf{49.7} & 62.34 \\
\bottomrule
\end{tabular}

\label{tab:thresholding_ablation}
\end{table}
Table~\ref{tab:thresholding_ablation} ablates the magnitude threshold $\tau$ used during post-processing. Increasing $\tau$ suppresses low-confidence regions and improves COCO$_S$ and COCO$_M$, with the best results at $\tau=150$. COCO$_L$ peaks at $\tau=100$, but the difference is small; therefore, we use $\tau=150$ as the default.

\subsection{Failure Cases}
\begin{figure*}[h]
    \centering

    \begin{minipage}[t]{0.23\textwidth}
        \centering
        \includegraphics[width=\linewidth]{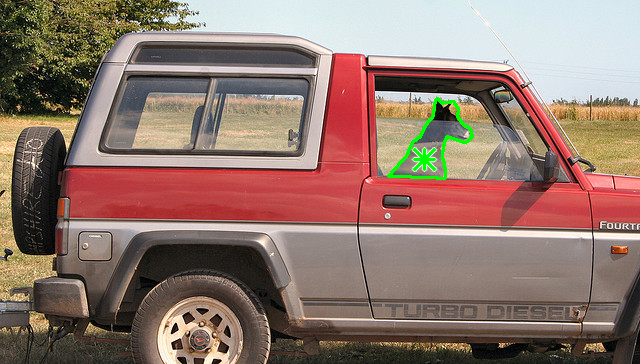}\\[-0.2em]
    \end{minipage}
    \hfill
    \begin{minipage}[t]{0.23\textwidth}
        \centering
        \includegraphics[width=\linewidth]{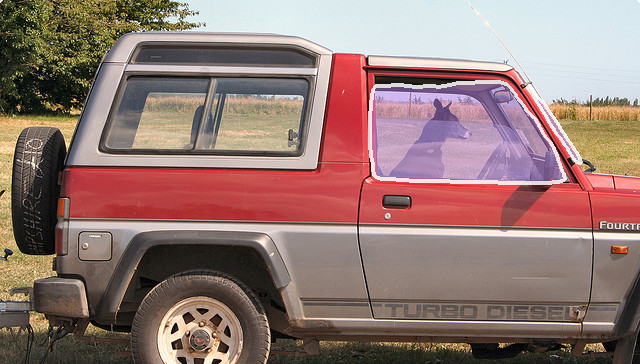}\\[-0.2em]
    \end{minipage}
    \hfill
    \begin{minipage}[t]{0.23\textwidth}
        \centering
        \includegraphics[width=\linewidth]{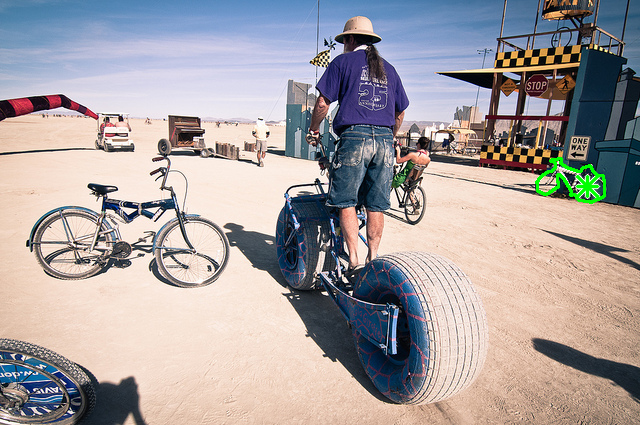}\\[-0.2em]
    \end{minipage}
    \hfill
    \begin{minipage}[t]{0.23\textwidth}
        \centering
        \includegraphics[width=\linewidth]{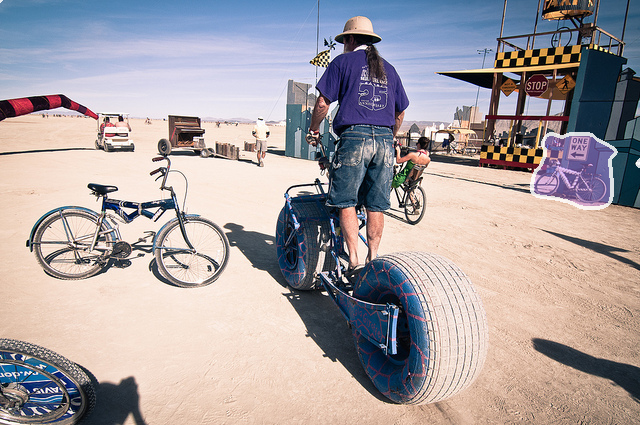}\\[-0.2em]
    \end{minipage}

    \vspace{0.5em}

    \begin{minipage}[t]{0.23\textwidth}
        \centering
        \includegraphics[width=\linewidth]{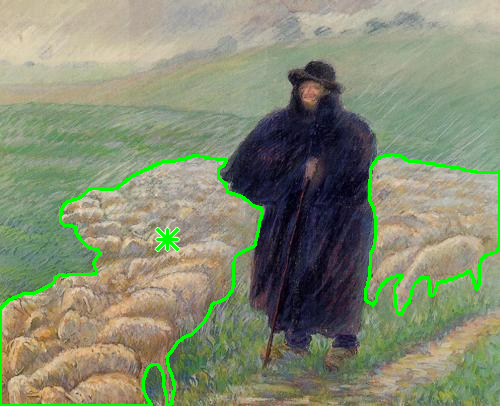}\\[-0.2em]
    \end{minipage}
    \hfill
    \begin{minipage}[t]{0.23\textwidth}
        \centering
        \includegraphics[width=\linewidth]{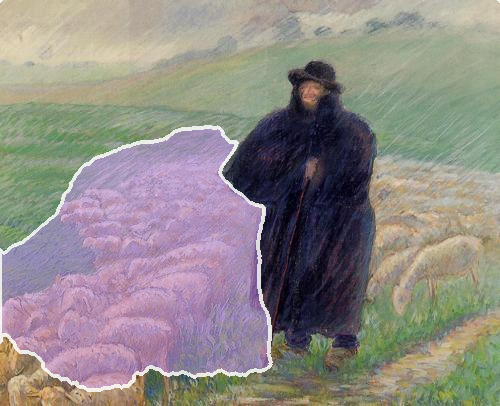}\\[-0.2em]
    \end{minipage}
    \hfill
    \begin{minipage}[t]{0.23\textwidth}
        \centering
        \includegraphics[width=\linewidth]{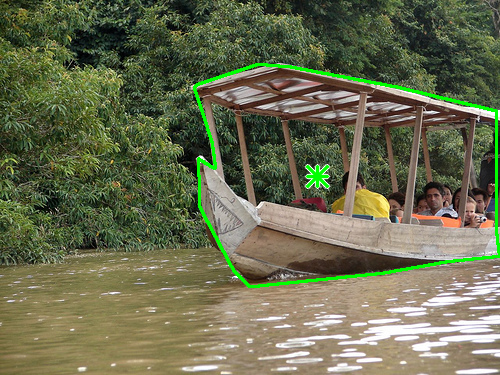}\\[-0.2em]
    \end{minipage}
    \hfill
    \begin{minipage}[t]{0.23\textwidth}
        \centering
        \includegraphics[width=\linewidth]{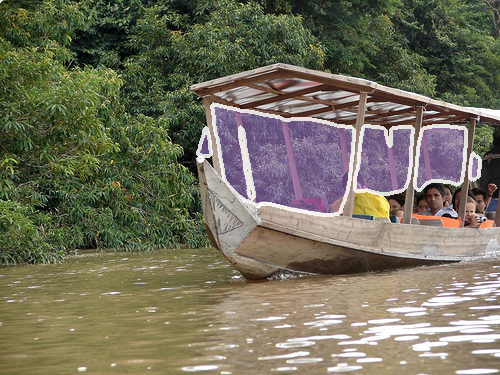}\\[-0.2em]
    \end{minipage}

\caption{Additional qualitative failure cases. For each example, the left image shows the input with the point prompt and target boundary overlay, while the right image shows the predicted output. The model captures the approximate target region but fails to precisely follow object boundaries or separates only part of the intended instance.}
    \label{fig:supp_failure_cases}
\end{figure*}
Figure~\ref{fig:supp_failure_cases} shows failure cases where our model misunderstood the prompt and either segmented only part of the object of interest or a different object. Additional prompts could help.


\newpage

\end{document}